\documentclass[conference]{IEEEtran}
\IEEEoverridecommandlockouts

\usepackage{cite}
\usepackage{amsmath,amssymb,amsfonts}
\usepackage{algorithmic}
\usepackage{graphicx}
\usepackage{subcaption}
\usepackage{textcomp}
\usepackage{comment}
\usepackage[none]{hyphenat}
\usepackage{xcolor}
\usepackage{eso-pic}
\def\BibTeX{{\rm B\kern-.05em{\sc i\kern-.025em b}\kern-.08em
    T\kern-.1667em\lower.7ex\hbox{E}\kern-.125emX}}

\begin{document} 

\makeatletter
\def\ps@IEEEtitlepagestyle{%
  \def\@oddfoot{\mycopyrightnotice}%
  \def\@evenfoot{}%
}
\def\mycopyrightnotice{%
  {\footnotesize  \hfill} 
}
\makeatother

\title{Low-Latency Relay Selection in NR-V2X Vehicular Communications via Graph Isomorphism Networks with Edge Features\\
\thanks{This work was supported by the European Union under the Italian National
Recovery and Resilience Plan (NRRP) of NextGenerationEU, partnership on
"Telecommunications of the Future" (PE00000001 - program "RESTART").}
}
\author{
\IEEEauthorblockN{ Giambattista Amati}
\IEEEauthorblockA{\textit{Fondazione Ugo Bordoni (FUB)}\\
Rome, Italy\\
gamati@fub.it}
\and
\IEEEauthorblockN{ Federica Mangiatordi}
\IEEEauthorblockA{\textit{Fondazione Ugo Bordoni (FUB)}\\
Rome, Italy\\
fmangiatordi@fub.it}
\and
\IEEEauthorblockN{ Emiliano Pallotti}
\IEEEauthorblockA{\textit{Fondazione Ugo Bordoni (FUB)}\\
Rome, Italy\\
epallotti@fub.it}
\and
\IEEEauthorblockN{ Simone Angelini}
\IEEEauthorblockA{\textit{Fondazione Ugo Bordoni (FUB)}\\
Rome, Italy\\
sangelini@fub.it}
\and
\IEEEauthorblockN{ Pierpaolo Salvo}
\IEEEauthorblockA{\textit{Fondazione Ugo Bordoni (FUB)}\\
Rome, Italy\\
psalvo@fub.it}
\and
\IEEEauthorblockN{ Paola Vocca}
\IEEEauthorblockA{\textit{University of Rome Tor Vergata (UniRoma2)}\\
Rome, Italy\\
paola.vocca@uniroma2.it}
}

\maketitle

\begin{abstract}
Reliable, low-latency uplink connectivity is a key requirement for 
C-V2X networks in dense urban environments, where fast channel variations and blockages often degrade direct vehicle-to-infrastructure links. Multi-hop relaying can restore coverage, but relay-link activation under radio, capacity, and routing constraints results in an NP-hard optimisation problem, typically solved via Mixed-Integer Linear Programming (MILP), whose runtime scales poorly with graph size.
This paper introduces an edge-aware Learning-to-Optimise framework for real-time relay selection. Each V2X  snapshot is modelled as a directed graph: node features encode vehicle state and traffic demand, while edge features capture radio-link capacity. An offline MILP oracle generates optimal relay configurations that supervise a Graph Isomorphism Network with Edge Features (GINE), enabling edge-level relay activation through a single forward pass, with tightly bounded inference latency. To bridge learning and exact optimisation, we also propose a hybrid GINE-Pruned MILP (GP-MILP) strategy in which GINE predictions prune the MILP search space. Experiments on a large-scale dataset generated via an OSM-SUMO-GEMV$^2$ pipeline show that GINE closely matches MILP decisions at the link level (accuracy 0.9589), F1-score (0.9544) on validation) and yields consistent end-to-end connectivity gains over a 1-hop MILP baseline (up to 9.2\% with four RSUs and 12\% with two RSUs). Inference latency remains tightly bounded, with all evaluated instances completing within 5~ms. Moreover, GP-MILP preserves MILP-equivalent solutions (same objective value) while achieving solver runtimes below 30~ms for more than 98\%) of the graph instances, making MILP-grade optimisation compatible with stringent NR-V2X latency budgets.
\end{abstract}


\section{Introduction}
Modern Connected and Automated Vehicles (CAVs) rely on
advanced onboard sensing technologies, such as Light Detection and Ranging (LiDAR), radar,
and high-resolution cameras, which continuously generate
large volumes of data. Supporting these data streams requires
robust and dependable uplink connectivity towards roadside
and core network infrastructures.
Next-generation 5G-Advanced and forthcoming 6G  NR-V2X
architectures significantly outperform legacy Dedicated Short-Range Communications (DSRC) and early  C-V2X solutions in terms of bandwidth availability and radio
resource management, enabling advanced V2X services with
stringent throughput, latency, and reliability requirements for
cooperative and autonomous driving
\cite{etsi_ts_122186_v18,3gpp_tr_23700_98,HAKAK2023100551,1010162021}.

Despite these technological advances, vehicular wireless
communications remain intrinsically exposed to severe channel
variability, frequent obstructions, and rapidly changing
propagation conditions, particularly in dense urban
environments \cite{Jameel2018PropagationCF,taiya2026ultra,clancy2024wireless}.
To mitigate these impairments, relay-assisted communication
has emerged as a key enabler for maintaining a reliable uplink
connectivity. Leveraging relaying through neighbouring CAVs,
 RSUs, or intelligent infrastructure elements
can effectively extend coverage and alleviate blockage effects
\cite{10304089}. However, the effectiveness of relay-based
solutions critically depend on the ability to select suitable
relay links under tight latency constraints and incomplete
network observability.

Relay selection in NR-V2X networks entails the joint
consideration of radio-layer feasibility, quality-of-service
(QoS) requirements, RSU capacity limitations, and multi-hop
flow conservation, while ensuring that each CAV maintains at
least one valid uplink path to the infrastructure. This coupling
of routing and resource constraints leads to an NP-hard
optimisation problem.  MILP formulations can compute  globally optimal relay
configurations, but their computational complexity increases
superlinearly with network size and graph density, rendering
them unsuitable for real-time decision-making in large-scale
urban scenarios \cite{Chen2024MILP}.

To address these limitations, this work adopts a
\emph{Learning-to-Optimise} (L2O) paradigm, in which neural
models are trained offline to approximate the structure of
optimal solutions and replace expensive optimisation routines
with fast, feed-forward inference at runtime
\cite{shao2021learningrobustcombinatorialoptimization,bengio2021machine}.
Graph Neural Networks (GNNs) are particularly well-suited to
this task, as NR-V2X environments can be naturally modelled
as graphs whose nodes represent CAVs and RSUs, and whose
edges capture time-varying wireless links
\cite{AmatiAEIT2025,10274765,11268640}. Through message
passing, GNNs can implicitly capture spatial, topological, and
radio-aware dependencies that are difficult to encode using
hand-crafted features.

In this paper, we formulate multi-hop relay selection in
NR-V2X networks as an \emph{edge-level learning-to-optimise
problem}, where the objective is to infer relay-link activation
decisions that maximise end-to-end uplink connectivity under
implicit global constraints. An edge-aware L2O framework
based on a \textit{Graph Isomorphism Network with Edge
Features} (GINE) is proposed, explicitly integrating radio-layer
link attributes into the message-passing process to guide relay
activation decisions. A MILP solver is employed offline as an
oracle to generate optimal relay configurations, which are then
used to supervise the training of the GINE model.

While the proposed model does not explicitly enforce routing
or flow-conservation constraints at inference time, empirical
results show that it closely approximates the MILP-optimal
solutions in terms of connectivity, while exhibiting stable and low inference latency across a wide range of network sizes.
Furthermore, the learned model is embedded within a hybrid
MILP+L2O pipeline, where GINE predictions are used to prune
the optimisation space of the MILP solver. We refer to this
hybrid strategy as \emph{GINE-Pruned MILP (GP-MILP)}.
This hybrid approach preserves MILP-grade optimality in practice while
reducing optimisation latency by orders of magnitude, making
exact optimisation viable within realistic NR-V2X latency
budgets.


\section{Related Work}
\label{sec:related_work}
Relay-assisted communication has been extensively studied to enhance uplink reliability and coverage in vehicular networks, particularly
under dense urban conditions characterised by blockage and fast channel
variations\cite{10945797}.
Early works addressed relay selection through heuristic or optimisation-based approaches, including shortest-path routing, greedy strategies, and MILP formulations that jointly consider radio
constraints, flow conservation, and infrastructure capacity.
While these methods can achieve optimal or near-optimal solutions, their computational complexity grows rapidly with network size and graph density, making them unsuitable for real-time NR-V2X operation \cite{Chen2024MILP}.

To overcome these limitations, learning-based approaches have recently been proposed for resource management and connectivity optimisation in V2X systems.
GNN have emerged as a natural modelling tool, as vehicular
networks can be represented as graphs with vehicles and infrastructure nodes connected by time-varying wireless links.
Existing GNN-based solutions have been applied to power control, spectrum
allocation, and link scheduling, demonstrating improved scalability compared to
traditional optimisation techniques \cite{chen2024gnn,10274765}. However, most of these methods do not explicitly account for global constraints induced by multi-hop routing and flow conservation.

More recent studies have explored edge-centric graph representations, where relay or link activation decisions are formulated as edge classification
problems or are addressed through line-graph constructions
\cite{jiang2020coembeddingnodesedgesgraph,AmatiAEIT2024,10068338}.
Although these approaches better capture link-level dependencies, they usually incur increased computational and memory overhead, and still do not enforce
global constraints such as multi-hop flow conservation or loop-free routing.

In contrast to prior work, this paper proposes an edge-aware
Learning-to-Optimise framework based on GINE, explicitly supervised by a MILP oracle.
The MILP formulation captures global optimality under NR-V2X constraints and is used exclusively offline to generate ground-truth relay configurations, while
the trained GINE model enables low-latency, near-constant-time inference at runtime.
To the best of our knowledge, this is among the first works that combine MILP-level optimality with edge-aware GNN learning to enable scalable, real-time multi-hop relay selection in NR-V2X networks. 





\section{System Model and Problem Statement}
\label{sec:system_model}
We consider an NR-V2X vehicular network operating within a bounded urban area,
composed of a set of CAVs and a set of RSUs connected to the core network.
The instantaneous network topology is modelled as a directed graph
$\mathcal{G} = (\mathcal{V}, \mathcal{E})$, where the node set
$\mathcal{V} = \mathcal{C} \cup \mathcal{R}$ includes both CAVs and RSUs, and each
directed edge $(u,v) \in \mathcal{E}$ represents a feasible NR-V2X wireless link
from node $u$ to node $v$ under radio and distance constraints.

Each node is characterised by its geographical position and functional role.
CAVs may generate uplink traffic associated with delay-sensitive V2X services,
while RSUs act as fixed infrastructure gateways and represent sink nodes for
uplink communication.
Wireless links are characterised by radio-layer attributes, including an
achievable capacity computed from instantaneous signal-to-noise ratio (SNR)
measurements using a Shannon-based model.

At each decision epoch, a CAV may either transmit its traffic directly to an RSU
via a cellular uplink or forward it through one or more intermediate CAVs using
NR-V2X sidelink communications.
To limit signalling overhead and interference, each CAV is constrained to
activate at most one outgoing relay link at a time, resulting in a single-path
uplink routing structure.
RSUs do not relay traffic and only act as termination points.

The relay selection problem involves selecting a subset of directed links to activate so that the largest possible number of CAVs can form a valid uplink route to at least one RSU, while ensuring the resulting paths are loop-free.
Due to the strong interdependence among relay
decisions and the combinatorial nature of path selection, the
problem is NP-hard.
MILP formulations can yield optimal solutions by jointly enforcing radio, flow, and topological constraints. However, their computational complexity grows rapidly with network size, which limits their practicality for making real-time decisions, such as selecting a relay link for a single connected vehicle (CAV) in dynamic urban NR-V2X scenarios.
This motivates the learning-based surrogate optimisation framework proposed in this paper.

\section{MILP Oracle for Multi-Hop Relay Selection}
\label{sec:milp_oracle}

To generate ground-truth relay configurations, we formulate the multi-hop uplink relay selection problem as a Mixed-Integer Linear Program (MILP) for each V2X graph modelling a V2X scenario.
The MILP captures global routing optimality under NR-V2X radio, flow, and capacity constraints and is used exclusively as an \emph{offline oracle} to
supervise the proposed Learning-to-Optimise framework.
Due to its combinatorial complexity, the MILP is not intended for real-time deployment.

\subsection{Decision Variables}
The optimisation uses the following variables:
\begin{itemize}[]
\item $v_{ij}\!\in\!\{0,1\}$: activation of link $(i,j)$;
\item $f_{ij}\!\ge\!0$: flow on link $(i,j)$;
\item $z_i\!\in\!\{0,1\}$: CAV $i$ connected to at least one RSU;
\item $u_i\!\in\!\{0,\dots,M\!-\!1\}$: ordering variable for cycle elimination.
\end{itemize}

\subsection{Objective Function}

The objective is to maximise the number of CAVs that establish a valid uplink
path to the infrastructure, possibly through multi-hop relaying:
\begin{equation}
    \max \sum_{i \in \mathrm{CAV}} z_i
\end{equation}

\subsection{Constraints}

\noindent\textbf{Flow conservation:}
For each CAV, the injected uplink demand must be conserved along the selected
multi-hop path:
\begin{equation}
\sum_{j} f_{ij} - \sum_{k} f_{ki} = d_i z_i,
\quad \forall i \in \mathrm{CAV}
\end{equation}
where $d_i$ denotes the traffic demand of CAV $i$.

\medskip
\noindent\textbf{Link and RSU capacity constraints:}
\begin{equation}
\begin{aligned}
f_{ij} &\le C_{ij}v_{ij}, &&\forall(i,j)\in\mathcal{E},\\
\sum_i f_{ij} &\le M_{\mathrm{RSU}_j}, &&\forall j\in\mathrm{RSU}.
\end{aligned}
\end{equation}

where $C_{ij}$ is the achievable link capacity derived from the instantaneous
radio conditions, and $M_{\mathrm{RSU}_j}$ denotes the maximum available capacity of RSU $j$.

\medskip
\noindent\textbf{Single-outgoing-link constraint:}
Each CAV can forward its traffic through at most one outgoing link:
\begin{equation}
\sum_{j:(i,j)\in\mathcal{E}} v_{ij} = z_i,
\quad \forall i \in \mathrm{CAV}
\end{equation}
\medskip
\noindent\textbf{Loop-free routing:}
To prevent routing cycles among relaying CAVs, Miller--Tucker--Zemlin (MTZ)
constraints are imposed on CAV--CAV links:
\begin{equation}
u_i + 1 \le u_j + M(1 - v_{ij}),
\quad \forall (i,j) \in \mathcal{E}_{\mathrm{CAV\text{-}CAV}}
\end{equation}

\subsection{Oracle Output}

Solving the MILP yields an optimal binary relay-activation matrix
$\mathbf{A}^\star = \{v_{ij}^\star\}$, where $v_{ij}^\star = 1$ indicates that
link $(i,j)$ belongs to an optimal multi-hop uplink configuration.
This matrix is used as edge-level supervision for training the proposed GINE-based Learning-to-Optimise model. By learning to approximate these MILP-optimal decisions, the GINE model provides
low-latency relay selection during inference, removing the need to solve the MILP.
The MILP formulation is run entirely offline during the dataset generation stage and has no effect on runtime performance.

\section{Edge-Aware Graph Learning for Low-Latency Relay Selection}
\label{sec:gine_l2o}
This section presents an edge-aware Learning-to-Optimise (L2O) framework for
low-latency multi-hop relay selection in NR-V2X networks.
The objective is to approximate the globally optimal relay configurations
computed by a MILP oracle through a single forward pass of a
GINE, enabling real-time inference
under stringent latency constraints.

\subsection{Graph Representation and Features}
\label{subsec:graph_features}
Each NR-V2X snapshot is represented as a directed graph
$\mathcal{G}=(\mathcal{V},\mathcal{E})$, where nodes correspond to CAVs and RSUs,
and directed edges denote feasible V2I or V2V wireless links.
Node features encode spatial position, node type, and uplink traffic demand, while each edge $(i,j)$ is associated with radio-aware features $\mathbf{e}_{ij}$, e.g. a feature
representing the achievable link capacity derived from instantaneous SNR.
This representation naturally captures the asymmetry and directionality of
uplink relay paths.

\begin{figure*}[t]
    \centering
    \includegraphics[width=0.6\linewidth]{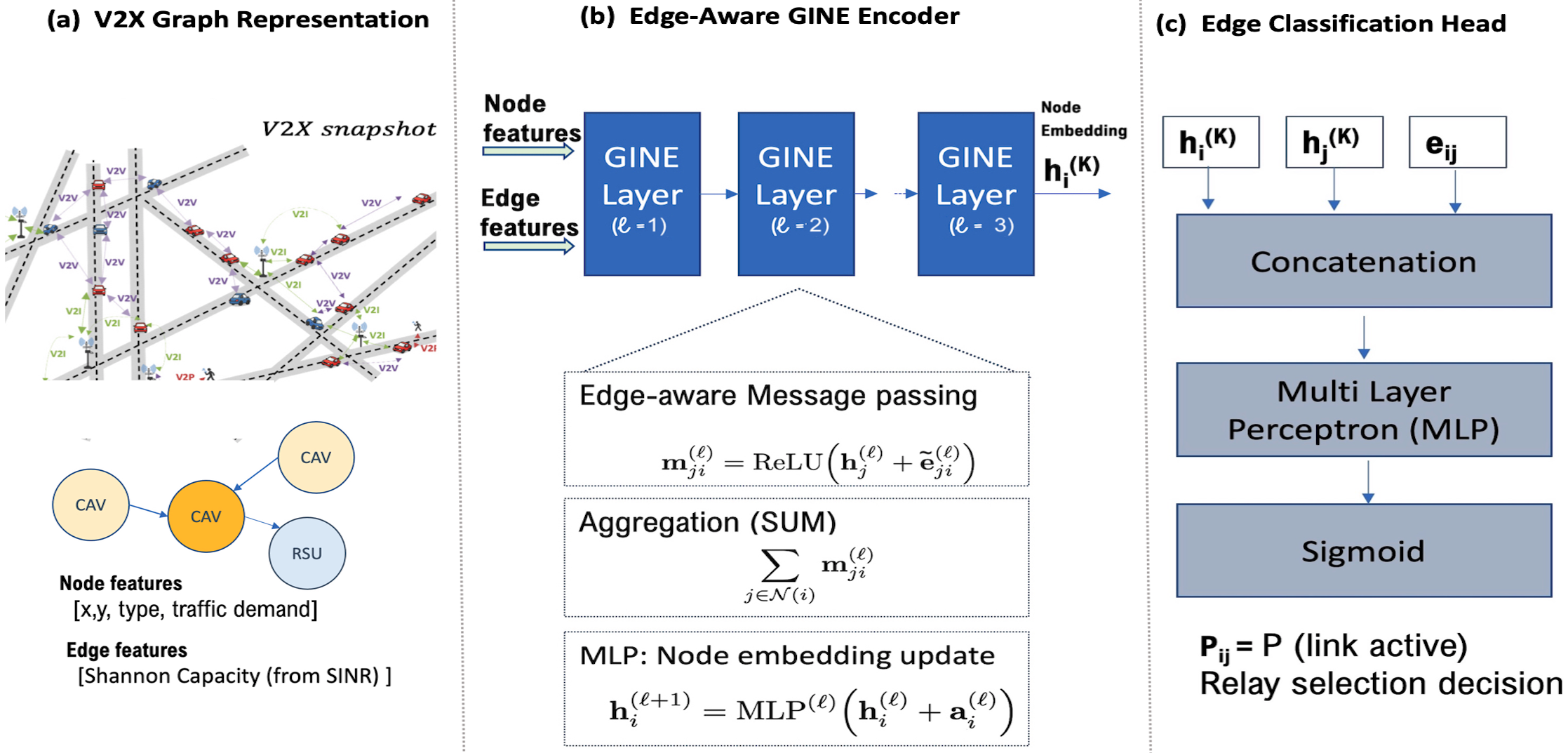}
    \caption{ GINE-based learning-to-optimise framework for
    multi-hop relay selection in NR-V2X networks.}
    \label{fig:gine_architecture}
\end{figure*}

\subsection{GINE Encoder}
\label{subsec:gine_encoder}
To jointly capture topological structure and radio-layer effects, a multi-layer
GINE is employed.
Unlike standard GIN architectures, GINE explicitly incorporates edge attributes
into the message-passing process, which is essential in NR-V2X scenarios where
link feasibility is strongly determined by radio conditions.
Given node embeddings $\mathbf{h}_i^{(\ell)}$, $\ell=1,\dots,L$, node updates are
\begin{equation}
\mathbf{h}_i^{(\ell+1)}=\mathrm{MLP}^{(\ell)}\!\Big(\mathbf{h}_i^{(\ell)}+\!\!\sum_{j\in\mathcal{N}(i)}\!\mathrm{ReLU}(\mathbf{h}_j^{(\ell)}+\tilde{\mathbf{e}}_{ij}^{(\ell)})\Big),
\end{equation}
where $\tilde{\mathbf{e}}_{ij}^{(\ell)}=\text{EdMLP}^{(\ell)}(\mathbf{e}_{ij})$ is a learned projection of the edge feature.  
EdMLP is a two-layer feed-forward network,
\begin{equation}
\text{EdMLP}(\mathbf{e}_{ij})
=\mathbf{W}_{e2}\,\text{Dropout}(\text{ReLU}(\mathbf{W}_{e1}\mathbf{e}_{ij}+\mathbf{b}_{e1}),p)+\mathbf{b}_{e2},
\end{equation}
with $\mathbf{e}_{ij}$ denoting raw edge features. Stacking $L=3$ GINE layers enables
the encoder to capture higher-order multi-hop dependencies that are critical for inferring structured relay paths.

\subsection{Edge Classification Head}
\label{subsec:edge_classifier}

Relay selection is formulated as a binary edge classification task.
For each directed edge $(i,j)$, a joint representation is constructed by
concatenating the embeddings of the incident nodes with the corresponding edge
feature:
\begin{equation}
\mathbf{z}_{ij} =
\left[
\mathbf{h}_i^{(L)} \,\|\, \mathbf{h}_j^{(L)} \,\|\, \mathbf{e}_{ij}
\right]
\end{equation}

This representation allows the classifier to jointly reason over node roles,
topological context, and radio quality.
A shared MLP produces an activation probability $p_{ij}$ indicating whether
edge $(i,j)$ should be selected as part of the relay configuration.

\subsection{Training Objective}
\label{subsec:training_objective}
The model is trained in a supervised manner using MILP-generated relay
configurations as ground truth.
Due to the strong imbalance between active and inactive edges, a class-weighted
binary cross-entropy loss is adopted:
\begin{equation}
\mathcal{L}_{\mathrm{BCE}}
=
-\!\!\sum_{(i,j)\in\mathcal{E}}
\!\!\!
\left[
\alpha\, y_{ij}\log p_{ij}
+
(1-y_{ij})\log(1-p_{ij})
\right]
\end{equation}
where $\alpha$ is chosen to penalise false negatives more strongly, as missing
relay links may break multi-hop connectivity.
Concretely, $\alpha$ is computed from the training set statistics as
$\alpha = \frac{N_{\mathrm{neg}}}{N_{\mathrm{pos}}}$
where $N_{\mathrm{pos}}$ and $N_{\mathrm{neg}}$ denote the number of active and
inactive edges in the training set, respectively.
This choice ensures balanced gradient contributions from positive and negative
samples and aligns the training objective with the connectivity-preservation
goal of the relay selection problem.
At validation time, a threshold sweep is performed to analyse the
precision--recall trade-off and adapt the relay selection policy to different
service requirements.

\section{Experimental Results}
\label{sec:ex_results}
This section evaluates the proposed GINE-based Learning-to-Optimise (L2O) framework with respect to (i) link-level agreement with the MILP oracle, (ii) end-to-end connectivity gains, and (iii) computational latency and scalability. 
Results are compared against the MILP oracle and a 1-hop MILP baseline to assess both optimality and the net gain enabled by multi-hop relaying.

\subsection{Experimental Setup and Dataset}
\label{subsec:exp_setup}

Experiments are conducted on a large-scale, realistic NR-V2X dataset generated through an integrated OSM--SUMO--GEMV$^2$ simulation pipeline  \cite{openstreetmap, SUMO2, gemv2website}. Vehicular mobility is modelled in SUMO over a $1\,\text{km}^2$ dense urban area of Rome (Porta Pia), preserving real road topology and traffic constraints, and yielding heterogeneous, time-varying vehicular graphs.
Radio propagation is simulated using the geometry-aware GEMV$^2$ model \cite{GEMV2}, compliant with 3GPP NR-V2X assumptions and accounting for distance-dependent path loss, LoS (Line of Sight)/NLoS conditions , and multipath effects.
Simulations use a 5.9~GHz carrier, 800~MHz effective bandwidth, RSU and vehicle transmit powers of 10~dBm and 0~dBm, respectively, and 1~dBi antenna gains. Each snapshot is represented as a directed graph containing both V2I and V2V links, whose edge features encode radio-aware metrics and Shannon-based capacities computed from the instantaneous SNR.
The number of active RSUs varies between 2 and 4, resulting in non-uniform infrastructure coverage, whereas CAVs continuously join and leave the region, yielding graphs of varying sizes and densities. Each snapshot is labelled offline using the MILP model in Section~\ref{sec:milp_oracle}, which determines the optimal multi-hop relay scheme subject to flow conservation, capacity, and loop-free routing constraints.
A total of $449{,}500$ graph snapshots are generated and split into training ($80\%$) and validation ($20\%$) sets. The GIN encoder consists of $L=3$ layers with 256-dimensional embeddings and is trained for 200 epochs with Adam, a learning rate of $10^{-3}$, a batch size of 64, and a dropout probability of 0.4. A class-weighted binary cross-entropy loss is employed to address the imbalance between active and iactive relay links \cite{Ghosh2024}.

\textbf{Runtime and latency measurement.}
Execution times are measured on a single server equipped with an Intel i9 CPU and an NVIDIA RTX 5000 GPU.
GINE inference is executed on the GPU in PyTorch \texttt{eval()} mode (no gradients) and includes the forward pass and probability thresholding.
To reduce timing jitter, we perform a warm-up of 10 runs and then report the average latency across 10 runs per graph instance (batch size = 1), excluding offline data-loading overhead.

MILP runtimes correspond to the end-to-end solver time per instance using Google OR-Tools, executed in multi-thread mode with a per-instance time limit of 11~s.

\subsection{Link-Level Prediction Performance}
\label{subsec:link_level_results}
We first assess the GINE-based L2O model's ability to approximate MILP decisions at the link level by formulating relay selection as a binary edge classification task.

At the nominal threshold, $ \tau = 0.5$, the model achieves an accuracy of $0.9589$, precision of $0.9508$, recall of $0.9581$, and F1-score of $0.9544$ on the validation set, indicating close agreement with MILP-optimal relay activations.
Comparable performance is observed on the full dataset, confirming stable generalisation across heterogeneous graph instances.
\begin{table}[h]
\centering
\caption{Link-level performance on the validation set at selected thresholds.}
\label{tab:link_level_thresholds}
\scriptsize
\setlength{\tabcolsep}{3pt}
\begin{tabular}{c c c c c}
\hline
\textbf{Thr.} & \textbf{Acc.} & \textbf{Prec.} & \textbf{Rec.} & \textbf{F1} \\
\hline
0.15 & 0.9251 & 0.8632 & 0.9902 & 0.9224 \\
0.45 & 0.9580 & 0.9449 & 0.9628 & 0.9538 \\
\textbf{0.50} & \textbf{0.9589} & \textbf{0.9508} & \textbf{0.9581} & \textbf{0.9544} \\
0.60 & 0.9580 & 0.9616 & 0.9463 & 0.9539 \\
0.75 & 0.9552 & 0.9726 & 0.9264 & 0.9489 \\
\hline
\end{tabular}
\end{table}

A threshold sweep highlights a controllable precision--recall trade-off. In recall-oriented regimes, lowering the threshold increases the probability of retaining MILP-active relay links, which is beneficial when preserving multi-hop connectivity is critical. Conversely, higher thresholds yield more conservative relay activation with increased precision.

\subsection{Connectivity Gain}
\label{subsec:connectivity_gain}

We evaluate end-to-end CAV connectivity as the fraction of vehicles that can
reach at least one RSU through the activated directed links, either via direct
V2I transmission or via multi-hop V2V relaying.

As a reference baseline, we consider a \emph{1-hop MILP} formulation obtained by
restricting the optimisation to direct CAV$\rightarrow$RSU links only, i.e.,
disabling CAV--CAV relaying. This baseline represents an oracle upper bound for
connectivity under \emph{no-relay} operation and provides a clean reference to
quantify the benefit introduced by multi-hop routing.

Compared to the 1-hop MILP baseline, enabling coordinated multi-hop relaying
yields consistent connectivity improvements. With four RSUs deployed, the
proposed framework achieves an average connectivity gain of $9.2\%$, while in
infrastructure-sparse scenarios with only two RSUs the gain increases up to
$12\%$. These results highlight the effectiveness of multi-hop relaying in
compensating for limited direct V2I coverage and underline the importance of
globally coordinated relay selection in dense urban NR-V2X environments.

\subsection{Inference Latency and Pruning-Assisted MILP}
At inference time, relay selection requires a single forward pass of the
GINE-based model followed by probability thresholding.
For a fixed number of message-passing layers, the computational complexity
scales linearly with the number of edges, i.e., $\mathcal{O}(|\mathcal{E}|)$,
in contrast to the exponential worst-case complexity of MILP optimisation.
This property enables predictable and low-latency decision-making, which is
essential for time-critical NR-V2X deployments.

\begin{figure}[th]
    \centering
    \includegraphics[width=0.95\linewidth]{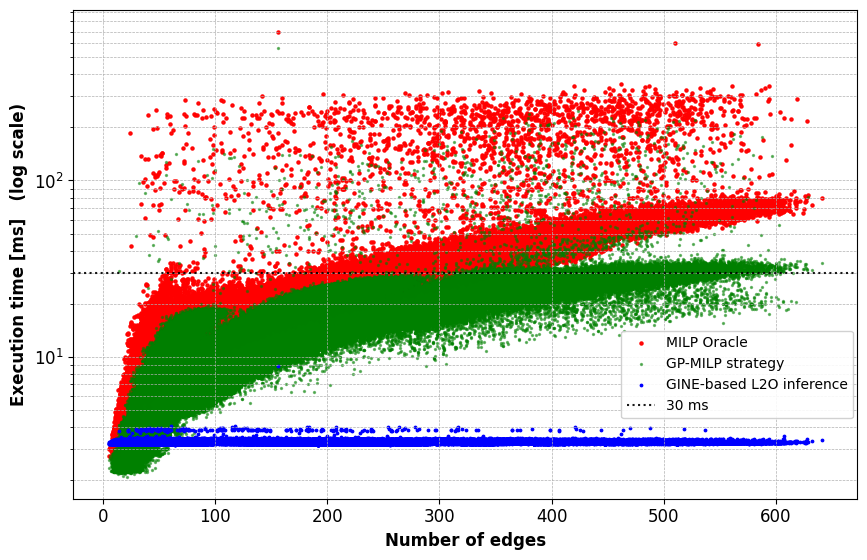}
    \caption{{Runtime as a function of the number of edges for the MILP oracle, the GINE-based L2O inference method, and the proposed hybrid \emph{GP-MILP} approach.}}
    \label{fig:exec-times}
\end{figure}

Figure~\ref{fig:exec-times} reports execution time as a function of graph size,
expressed by the number of edges. For readability, MILP runtimes above 9~s are clipped in the plot; all instances are included in the aggregate statistics reported in Table~\ref{tab:latency_thresholds_extended}.

The GINE-based model exhibits a highly concentrated inference-time distribution,
with execution times consistently within a few milliseconds and no observable
dependence on graph size or edge density within the evaluated range. This behaviour is consistent with the linear computational complexity of message
passing with respect to $|\mathcal{E}|$ for fixed network depth $L$. In
contrast, the MILP oracle shows rapidly increasing runtime as graph size grows,
reflecting the combinatorial nature of exact optimisation under multi-hop and
flow-conservation constraints.

We also evaluate a hybrid optimisation strategy, termed
\emph{GINE-Pruned MILP (GP-MILP)}, in which the MILP solver is applied to a
reduced graph obtained by pruning edges using GINE predictions. Pruning is
performed at a recall-oriented operating point ($Thr.=0.15$), to retain oracle-active links
with high probability ($99\%$), thereby reducing the MILP search space while preserving
solution quality.

Experimental results demonstrate that GP-MILP execution times stay below 30~ms for nearly all graph instances, while preserving MILP-equivalent solutions on the evaluated dataset, i.e., yielding the same objective value and connectivity outcome as the MILP solved on the complete graphs.

\begin{table}[h]
\centering
\caption{Percentage of graph instances meeting given latency thresholds for GINE-based L2O and MILP after edge pruning.}
\label{tab:latency_thresholds_extended}
\begin{tabular}{lccc}
\hline
\textbf{Method} & \textbf{$<5$ ms} & \textbf{$<10$ ms} & \textbf{$<30$ ms}  \\
\hline
GINE-based L2O & 100.0\% & 100.0\% & 100.0\% \\
GP-MILP & 10.15\% & 33.88\% & 98.14\% \\
MILP Oracle & 1.1\% & 12.44\% & 44.87\% \\
\hline
\end{tabular}
\end{table}

Taken together, these findings show that the proposed GINE-based L2O framework supports reliable, millisecond-level relay selection and that the pruning-enhanced pipeline renders GP-MILP  optimisation stategy effectively compatible with strict NR-V2X latency constraints.
\label{subsec:latency}

\section{Conclusions}
\label{sec:conclusion}

This paper investigated low-latency multi-hop relay selection in NR-V2X
networks under realistic urban propagation conditions. We proposed an
edge-aware Learning-to-Optimise framework based on a  GINE, trained offline under MILP supervision to infer
relay-link activations that support end-to-end uplink connectivity.

By explicitly incorporating radio-layer link attributes into the message-passing
procedure, the proposed model captures both topological context and link quality
effects that are critical for multi-hop relaying. Experimental results on a
large-scale dataset generated via an OSM--SUMO--GEMV$^2$ pipeline show that the
GINE-based approach closely matches the MILP oracle in terms of link-level
agreement and achieves consistent connectivity improvements over a 1-hop MILP baseline, while exhibiting stable and low inference latency across the evaluated
range of graph sizes.
In addition, we evaluated the proposed hybrid \emph{GP-MILP} strategy in which GINE predictions
are used to prune the MILP solver's optimisation space. On the considered
dataset, this pruning-based pipeline attains MILP-equivalent solutions while
reducing solver runtime by orders of magnitude, making MILP-grade optimisation
practically compatible with stringent latency budgets.

Future work will focus on temporal graph sequences and constraint-aware post-processing to further enhance feasibility, as well as on validating the method in distributed RSU/MEC-assisted deployments and multi-service NR-V2X environments.


\bibliographystyle{IEEEtran}
\bibliography{IEEEabrv, References.bib}
\newpage

\end{document}